\documentclass[conference]{IEEEtran}
\IEEEoverridecommandlockouts
\usepackage{multirow}
\usepackage{multicol}
\usepackage[utf8]{inputenc} 
\usepackage{kotex} 

\usepackage{hyperref}
\usepackage{cite}
\usepackage{amsmath,amssymb,amsfonts}
\usepackage{algorithm}
\usepackage{algorithmicx}
\usepackage{algpseudocode}
\usepackage{graphicx}
\usepackage{textcomp}
\usepackage{xcolor}
\usepackage{xspace}
\usepackage{soul}
\usepackage{kotex}
\usepackage{booktabs} 
\usepackage{hyperref}
\definecolor{linkcolour}{rgb}{0,0.2,0.6}
\definecolor{xgreen}{rgb}{0.2,0.6,0.0}
\definecolor{xred}{rgb}{0.7,0.1,0.0}
\hypersetup{colorlinks,breaklinks,citecolor=red!50, urlcolor=linkcolour, linkcolor=xgreen}

\usepackage{amsmath}
\usepackage{color}
\usepackage{array}
\usepackage{stfloats}
\usepackage{amsthm} 
\usepackage{amssymb}
\usepackage{float}
\usepackage{nccmath}
\usepackage{graphicx}
\usepackage{setspace}
\usepackage[normalem]{ulem}
\usepackage{booktabs}
\usepackage{subcaption}
\usepackage{mwe}
\usepackage{cite}
\usepackage{graphicx}
\usepackage[export]{adjustbox}
\usepackage{amsmath,amssymb,amsfonts}
\usepackage{graphicx}
\usepackage{textcomp}
\usepackage{xcolor}
\usepackage[utf8]{inputenc} 
\usepackage{kotex} 

\usepackage{algorithm} 
\usepackage{algpseudocode}
\usepackage{grffile}
\usepackage{svg}

\usepackage{kotex}
\newcommand{\RN}[1]{%
  \textup{\uppercase\expandafter{\romannumeral#1}}%
}
\newcommand{\BfPara}[1]{\vspace{1mm}{\noindent\bf#1.}\xspace}

\def\BibTeX{{\rm B\kern-.05em{\sc i\kern-.025em b}\kern-.08em
    T\kern-.1667em\lower.7ex\hbox{E}\kern-.125emX}}
\begin{document}

\title{Search Space Adaptation for Differentiable Neural Architecture Search in Image Classification}

\author{
\IEEEauthorblockN{Youngkee Kim}
\IEEEauthorblockA{\textit{Korea University} \\
Seoul, Korea \\
\texttt{felixkim@korea.ac.kr}}
\and
\IEEEauthorblockN{Soyi Jung}
\IEEEauthorblockA{\textit{Hallym University} \\
Chuncheon, Korea \\
\texttt{sjung@hallym.ac.kr}}
\and
\IEEEauthorblockN{Minseok Choi}
\IEEEauthorblockA{\textit{Kyung Hee University} \\
Yong-in, Korea \\
\texttt{choims@khu.ac.kr}}
\and
\IEEEauthorblockN{Joongheon Kim}
\IEEEauthorblockA{\textit{Korea University}\\
Seoul, Korea \\
\texttt{joongheon@korea.ac.kr}}
}

\maketitle

\begin{abstract}
As deep neural networks achieve unprecedented performance in various tasks, neural architecture search (NAS), a research field for designing neural network architectures with automated processes, is actively underway. More recently, differentiable NAS has a great impact by reducing the search cost to the level of training a single network. Besides, the search space that defines candidate architectures to be searched directly affects the performance of the final architecture. In this paper, we propose an adaptation scheme of the search space by introducing a search scope. The effectiveness of proposed method is demonstrated with ProxylessNAS for the image classification task. Furthermore, we visualize the trajectory of architecture parameter updates and provide insights to improve the architecture search.
\end{abstract}

\section{Introduction}
In recent decades, deep neural networks (DNNs) in machine learning (ML) have excelled in computer vision tasks with the ability of convolution neural networks (CNNs) to extract spatial information~\cite{baek2021joint},~\cite{pieee202105park}. Designing an appropriate architecture of DNN for the target task is essential to obtain good performances; however, it is quite heuristic and unexplainable\cite{tvt201905shin}. For this reason, neural architecture search (NAS) has emerged to optimize the neural architecture design.

One of early NAS studies introduces a reinforcement learning (RL)-based controller that constructs candidate architectures by iteratively selecting architectural hyperparameters~\cite{zoph2017nas}. A subsequent study demonstrates that RL-based NAS can discover new architectures outperforming handcrafted architectures by human experts~\cite{zoph2018NasNet}. In addition, another NAS approach achieves cutting edge with an evolutionary algorithm that continues until the best architecture remains~\cite{amoeba2019real}.

Despite the accomplishment of conventional NAS, the iterative training for candidate architectures incurs enormous computational costs that hinder practical applications~\cite{kim2022two}. To cope with this problem, an one-shot NAS approach has emerged, which aims to train only one neural network instead of training each candidate independently~\cite{kim2021trends}. In particular, a study proposes a differentiable NAS (DNAS) approach that applies a continuous relaxation to candidate operations so that the search space can be represented as a differentiable overparameterized neural network~\cite{darts2018liu}.
Accordingly, many follow-up studies have been conducted based on DNAS due to the significant reduction in the search cost. However, DNAS approaches are GPU memory-intensive because the entire overparameterized network has to be loaded onto GPUs during search. For this reason, a technique for binarizing architecture parameters of the overparameterized network at run-time has been proposed to address the GPU memory problem~\cite{proxyless2019cai}.

A common fact of the above NAS approaches is that the search space is closely related to the performance of the final architecture. In other words, the search space has to be defined carefully, considering the trade-off between the search cost and the quality of the final architecture.

\section{ProxylessNAS}
ProxylessNAS is one of the DNAS-based approaches that aims to directly search architectures for large-scale target tasks rather than proxy tasks.~\cite{proxyless2019cai}. We modify and utilize the strategy of ProxylessNAS to search our proposed search space. 
Similar to typical DNAS approaches, ProxylessNAS alternately optimizes model weights and architecture parameters via gradient-based methods. However, ProxylessNAS introduces binary gates that binarize architecture parameters of an overparameterized network to load only one active path at run-time. This technique alleviates the high GPU memory consumption issue of DNAS caused by loading the entire overparameterized network to update model weights. Therefore, ProxylessNAS can search a large search space and dataset by reducing the memory footprint of the network to load onto GPUs.

Also, ProxylessNAS considers inference latency as the objective of its architecture search.
Most hardware-aware NAS approaches calculate FLOPs of candidate architectures to reflect the inference latency. On the other hand, ProxylessNAS computes the expected latency of candidate architectures based on the pre-calculated latency of each operation. The expected latency is formulated in a differentiable way and incorporated into the loss function. Note that we exclude the latency related term from our loss function to focus on the effect related to the accuracy.

\section{Search Space Adaptation}\label{sec:searchspace}
From the DNAS perspective, the search space is a set of candidate architectures and can be represented as an overparameterized network. Note that we refer to the overparameterized network as the search network in this section. Our proposed adaptive search space is defined by a search unit block and a search scope. The search unit block is determined by candidate operations, and the search scope indicates the coverage of the base architecture to be replaced by the search unit block.
\begin{figure}[t!]
\center
    \includegraphics[width=0.475\columnwidth]{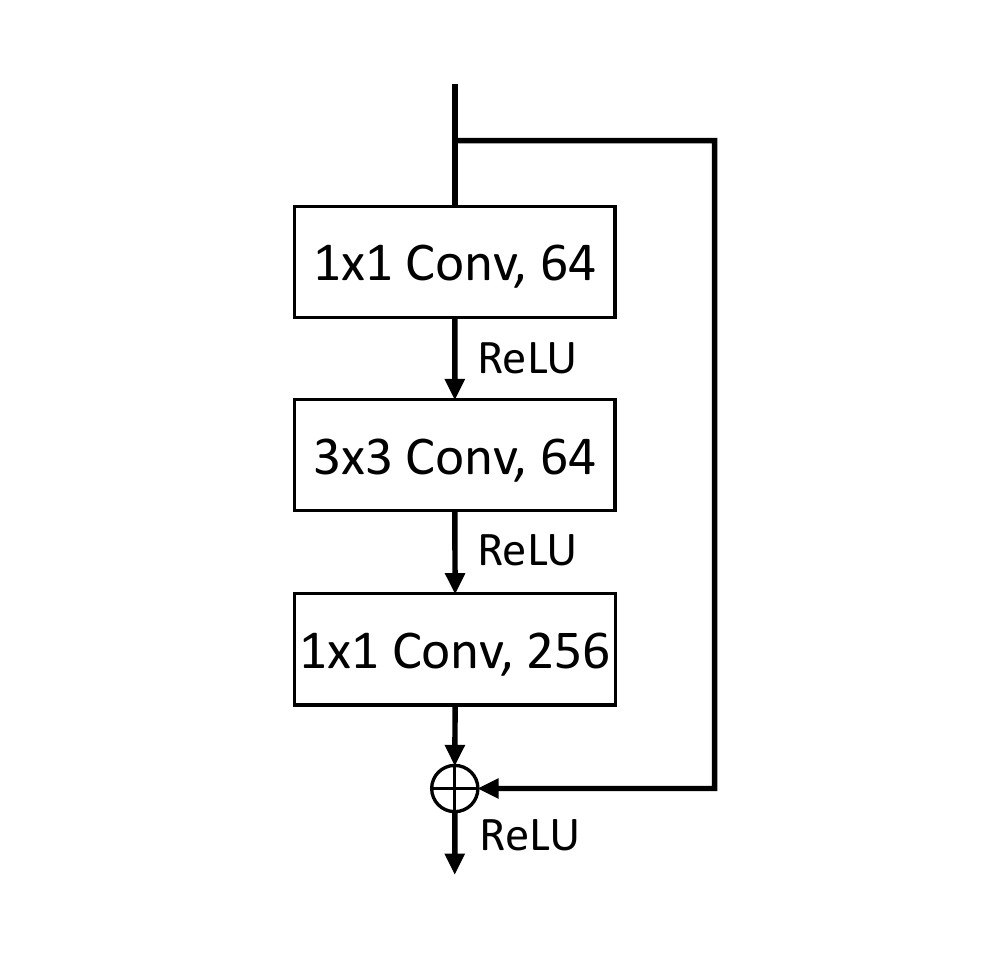}
    \includegraphics[width=0.475\columnwidth]{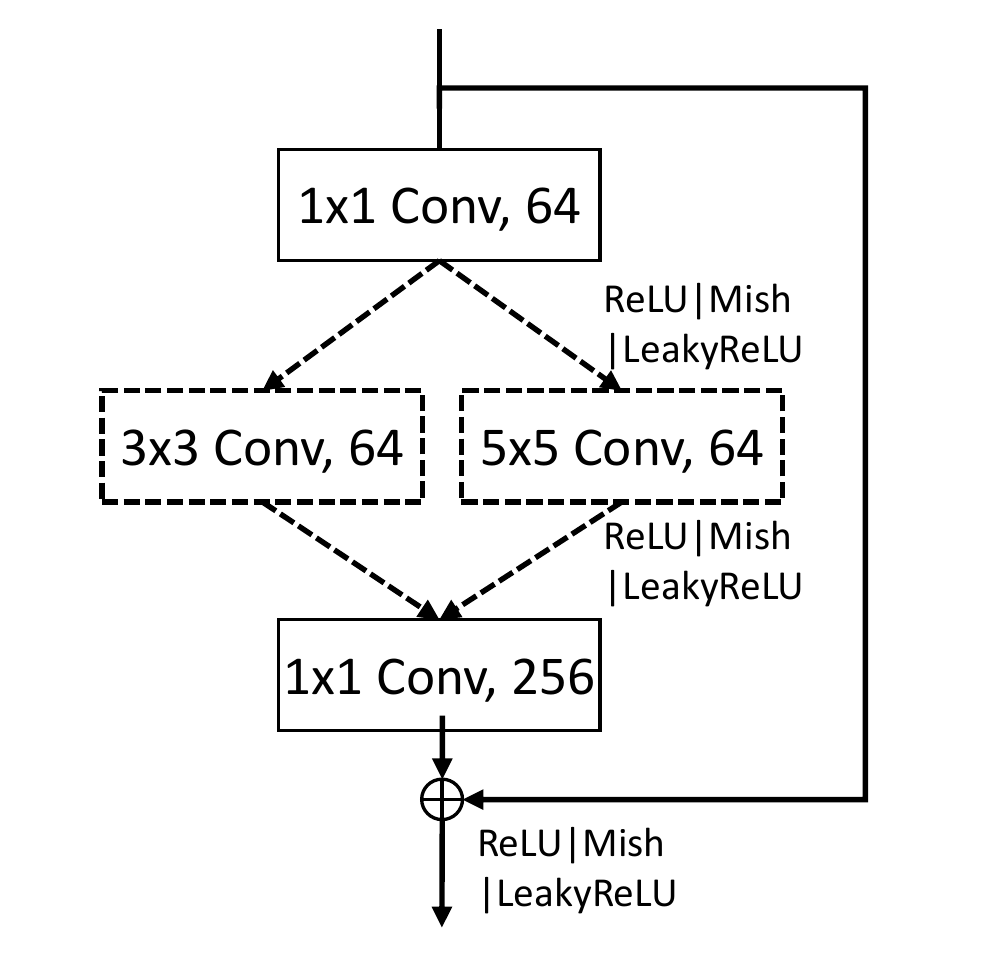}\\
    \centering (a) Bottleneck block.
    \qquad (b) Search unit block.\\
\caption{Unit blocks of ResNet-50 and search network.}
\label{fig:blocks}
\vspace{-2mm}
\end{figure}
\subsection{Search unit block}
The proposed search space refers to ResNet-50 as the base architecture~\cite{resnet2016he}. ResNet-50 consists of 16 bottleneck blocks, with a total of 50 layers. The bottleneck block and the search unit block are depicted in the Fig.~\ref{fig:blocks}. We define the search unit block by placing our candidate operations in the bottleneck block. There are two factors that define our candidate operations. One is the kernel size of convolution filter (i.e., 3, 5), and the other is the type of activation function (i.e., ReLU, LeakyReLU, Mish) following convolution operations. The variation of kernel size offers the possibility to improve the performance of CNNs by changing the size of receptive fields~\cite{receptive}. We also include a couple of recently used activation functions based on the previous research that the type of activation affects the accuracy of the same model~\cite{mish2020misra}.
\begin{figure}[t!]
\centering
\begin{tabular}{c}
\includegraphics[width=1 \columnwidth]{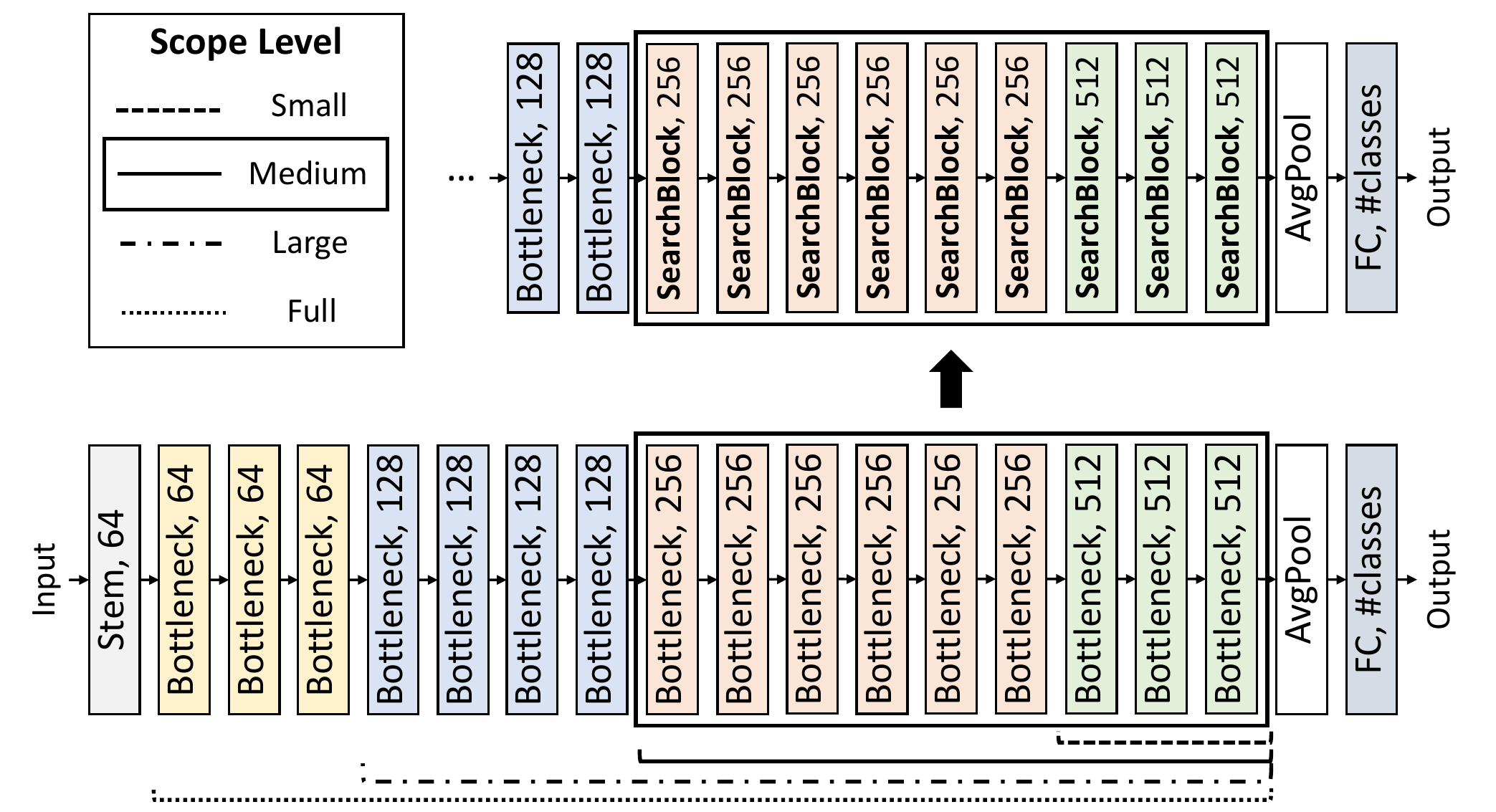}
\end{tabular}
\caption{Transition to search network by selected search scope.}
\label{fig:searchscope}
\vspace{-6mm}
\end{figure}
\subsection{Search scope}
The adaptive search space is determined by selecting a search scope level. The search network is constructed by replacing bottleneck blocks of ResNet-50 within the search scope with predefined search unit blocks. We first group the architecture of ResNet-50 into four parts according to the number of feature channels. Then we determine four levels of search scope starting from the deepest part and gradually expanding. The idea of deep-to-shallow expansion comes from the previous research that features eventually transit from general to specific along CNNs~\cite{last_k_finetuning}. Fig.~\ref{fig:searchscope} depicts the transition to the search network by the selected medium search scope. As a result, a wider search scope generates a larger search network, which means that more candidate architectures are considered during the search.

\section{Experiments and Results}
\subsection{Implementation details}
\BfPara{Dataset}
We demonstrate the proposed search space with ProxylessNAS on CIFAR-10 for the image classification task. In order to focus on the effectiveness of the search space, we apply common data preprocessing. The input image is cropped into random parts and resized to $224\times224$. The resized image is flipped horizontally with a given probability of 0.5 before normalization. The batch size for both training and validation data is set to 64.

\BfPara{Framework}
We leverage the Neural Network Intelligence (NNI) with Pytorch to implement our experiments. NNI is an open-source AutoML toolkit led by Microsoft that covers the entire ML pipeline, including feature engineering, neural architecture search, hyperparameter tuning and model compression~\cite{nni2022}. We demonstrate the proposed search space with customized ProxylessNAS implemented on NNI.

\BfPara{Training details}
Our DNAS with ProxylessNAS trains the search network via gradient-based methods. Therefore, there are two kinds of training, one is training of the search network and the other is retraining of the discovered architecture. In the search network training, model weights of the search network are updated on the training set using the SGD optimizer with Nesterov momentum. We set the initial learning rate to 0.05, the momentum value to 0.9 and the weight decay to 4e-5. Conversely, the Adam optimizer with an initial learning rate of 0.001 and no weight decay is used to update architecture parameters on the validation set. The path with the highest architecture parameters constitutes the final architecture after the search training. Finally, we retrain the determined architecture from scratch on the training set, under the same settings for updating model weights in the search training.

\subsection{Results}
As we explain in Sec.~\ref{sec:searchspace}, each level of search scope is applied to obtain the search space. We independently perform architecture searches for each search space and evaluate discovered architectures under the identical setting. Furthermore, we visualize the trajectory of architecture parameter updates and identify insights from trends.\\

\begin{table}[t!]
    \centering \normalsize
    \begin{tabular}{l|c|c}
    \toprule[1pt]
        \multicolumn{1}{c|}{\textbf{Model}} & \multicolumn{1}{c|}{\textbf{Params}} & \multicolumn{1}{c}{\textbf{Top-1 Accuracy (\%)}} \\\midrule
         ResNet-50 (baseline) & 23.53M & 91.10\\
         PResNet-50-s (ours) & 23.53M & 92.80\\
         PResNet-50-m (ours) & 24.58M & 92.35\\
         PResNet-50-l (ours) & 23.79M & 92.07\\
         PResNet-50-f (ours) & 25.23M & \textbf{93.22}\\\bottomrule
    \end{tabular}
    \vspace{-5pt}
    \caption{Performance comparison on CIFAR-10.}
    \label{tab:performance}
\vspace{-5mm}
\end{table}

\begin{figure}[t!]
\centering
\begin{tabular}{c}
\includegraphics[width=1 \columnwidth]{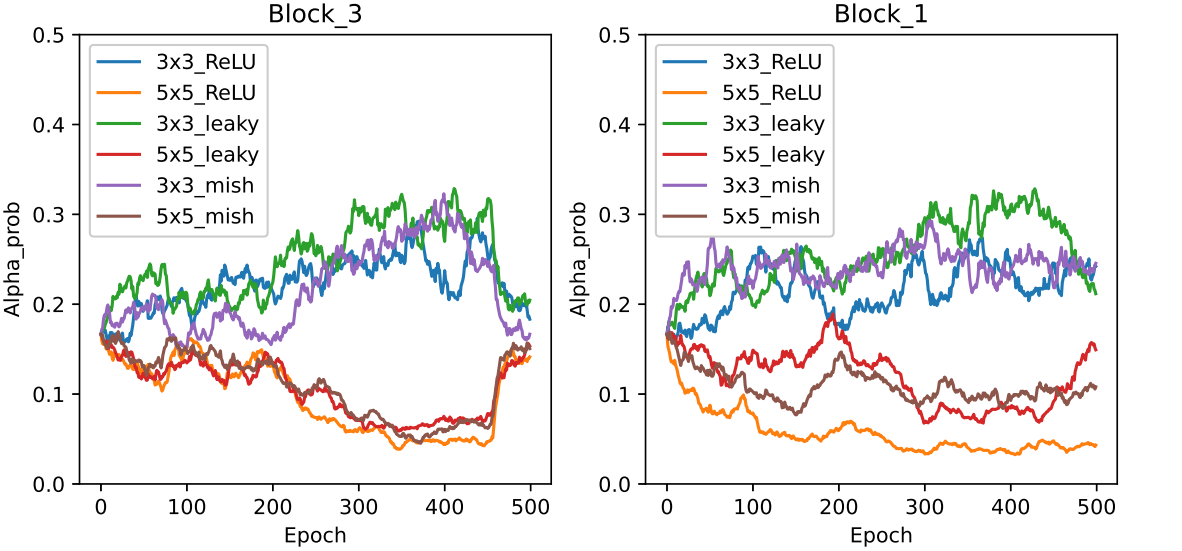}
\end{tabular}
\caption{Architecture parameter update trajectories with the large search scope.}
\label{fig:arch_probs}
\vspace{-7mm}
\end{figure}
\BfPara{Performance comparison of searched architectures}
We explore each search space based on aforementioned training details and evaluate the performance of searched architectures by retraining from scratch. Top-1 accuracy results of searched architectures on CIFAR-10 are summarized in Table~\ref{tab:performance}. We name the family of searched architectures as PResNet-50 with a letter indicating the level of search scope. Note that searched architectures outperform the baseline by up to 2.12\%. Some models have more parameters than the baseline due to the fact that we consider larger kernel size of convolution filter, while the type of activation function does not affect the size of parameters.

Although PResNet-50-f is superior to other searched models, it is the most expensive model in terms of search time due to the huge search space. The search training for PResNet-50-f consumes about $\times2.07$ longer than PResNet-50-s on a GeForce RTX 2080 Ti. Therefore, a smaller search scope can be a good choice depending on a given budget.

\BfPara{Visualization of architecture parameter updates}
We track changes in architecture parameters to capture properties of the proposed search space during the search. Fig.~\ref{fig:arch_probs} shows trajectories of architecture parameter updates of the deepest and third deepest search blocks with the large search scope. Block numbers are assigned sequentially from the deepest search block. Note that represented values on the y-axis are calculated by applying a softmax to real-valued architecture parameters for better understanding. 
Firstly, architecture parameters with a smaller search scope converge relatively quickly compared to a larger scope. In Fig.~\ref{fig:arch_probs}, the kernel size of both search blocks is determined to be 3 at approximately 300 search epochs, whereas it takes less than 100 epochs for the small search scope. The type of activation function seems meaningless on these blocks, but has a different effect on other blocks.
Secondly, we find that architecture parameters of some blocks tend to become unstable again when the search lasts excessively. In Fig.~\ref{fig:arch_probs}, architecture parameters are updated in a certain direction until 400 epochs, while they show erratic movements after that. Thus, a carefully designed criterion for determining when to stop the search is important to avoid performance degradation.

\section{Conclusion}
We introduce an adaptation of search space with a search unit block and selectable search scopes. The effectiveness of proposed method is verified with customized ProxylessNAS implemented on NNI framework. We also provide several insights to improve DNAS by tracking changes in architecture parameters during the search. Future work could be a study of a tailored search scope for a given circumstance and appropriate criteria for the search exit point.

\section*{Acknowledgment}
This work was supported by Institute of Information \& Communications Technology Planning \& Evaluation (IITP) grant funded by the Korea government(MSIT) (No. 2021-0-00766, Development of Integrated Development Framework that supports Automatic Neural Network Generation and Deployment optimized for Runtime Environment). Joongheon Kim is a corresponding author of this paper.

\bibliographystyle{IEEEtran}
\bibliography{ref_aimlab,ref}
\end{document}